\documentclass[letterpaper, 10 pt, conference]{ieeeconf}  

  \usepackage[pdftex]{graphicx}
   \DeclareGraphicsExtensions{.pdf,.jpeg,.png}

\IEEEoverridecommandlockouts                              

\title{\LARGE \bf
Vehicle Local Position Estimation System
}

\author{Mrinal Haloi$^{1}$ ,Dinesh Babu Jayagopi$^{2}$
\thanks{Accepted in ICVES-2014, hyderabad,India. Copyright 2015 by the authors. $^{1}$ Indian Institute of Technology, Guwahati, India.
        {\tt\small (h.mrinal@iitg.ernet.in or mrinal.haloi11@gmail.com)}}%
\thanks{$^{2}$International Institute of Information Technology, Bangalore
       {\tt\small jdinesh@iiitb.ac.in}}%
}

\begin{document}

\maketitle
\thispagestyle{empty}
\pagestyle{empty}

\begin{abstract}

In this paper, a robust vehicle local position estimation with the help of single camera sensor and GPS is presented. A modified Inverse Perspective Mapping, illuminant Invariant techniques and object detection based approach is used to localize the vehicle in the road. Vehicles current lane, its position from road boundary and other cars are used to define its local position. For this purpose Lane markings are detected using a Laplacian edge feature, robust to shadowing. Effect of shadowing and extra sun light are removed using Lab color space and illuminant invariant techniques. Lanes are assumed to be as parabolic model and fitted using robust RANSAC. This method can reliably detect all lanes of the road, estimate lane departure angle and local position of vehicle relative to lanes, road boundary and other cars. Different type of obstacle like pedestrians, vehicles are detected using HOG feature based deformable part model.

\end{abstract}

\section{INTRODUCTION}

With the increasing number of accident vital life loses,India is one of the most accident prone Country, where according to the NCRB report 135,000 died in 2013 and property damage of \$ 20 billion [22].Many a time, accidents
and unusual traffic congestion take place due to careless and impatient nature of drivers. In most cases drivers don't follow lane rules and traffic rules leading to traffic congestion and accidents. For countermeasuring all this problem we need advanced driver assistance system that can assist people drive safely or drive itself safely in case of autonomous driving cars. It always a challenge to make autonomous car that can self sense the environment and drive like a aware human. In some recent works researchers have developed autonomous car, eventhough its not still deployable in real life.

In developed countries like U.S.A,Germany, with the gradual emergence of autonomous driving research, efforts are on to build a smart driving system that can drive more safely without any fatigue, as compared to humans can be programmed to follow traffic rules. Main challenge involves understanding complex traffic patterns and taking real time decision on the basis of visual data from camera and laser sensor data etc. For these automatic cars and also to help human drivers, modelling vehicle local position with respect to the road environment is very relevant for accurate driving, maintaining correct lane and keeping track of front vehicles. 

While driving in a road it is very important to keep track of front vehilces, their relative speeds and current lane, this will help avoiding accident and undesireable traffic jam. If we lose our attention while driving in a busy road, we may cause accident, since our car is not equipied with new system that can warn us about various issues.
\begin{figure}[h!]
  \centering
      \includegraphics[width=3.2in,height=2.1in]{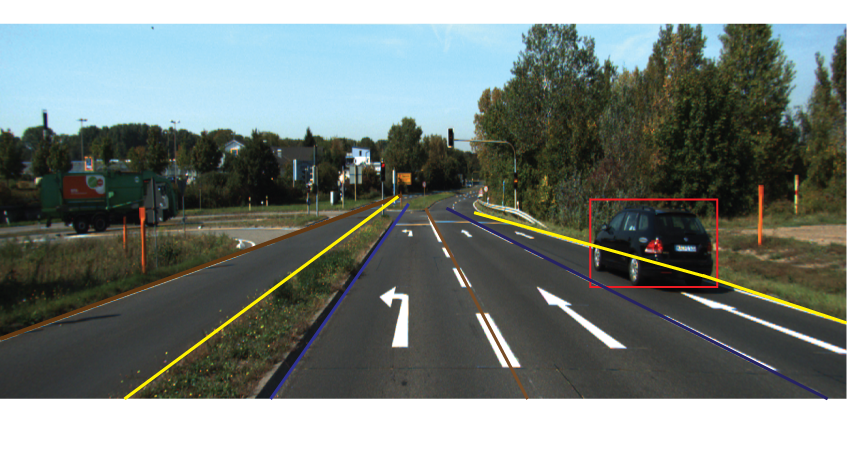}
\caption{Vehicle positioning scenarios}
\end{figure}

 In this work we have addressed various important problems in driving, specifically to enhance driver safety, developed a local vehicle position estimation system with respect to the road environment and combined this infomation with GPS data for getting precise global location for smart driving experience. This formulation will answer about those questions, like which lane i am in?, my position from road boundary?, others vehicles position with respect to me?, is other cars are moving? While driving in the car this information can help the driver for taking smarter and better action in real time to avoid accident. Using vehicle current lane information,other cars movement and its location from front and side vehicles on the road, we accurately modelled its position. Also a robust lane detection system is proposed and used for collecting lane information. Recent state of the art object detection method is used to localize and detect cars on the road. For data collecting we have used wide angle camera sensor to capture surrounding road environment and GPS sensor for global position data. 

Rest of the paper is organised as follows, in section 2 we have described related literature, localisation of vehicle with respect to lane ans other cars passing by are elaborated in section 3 and 4 and section 5 describes experiment setup and result obtained.

\section{Related Work}
The related literature on autonomous driving and advanced driving assistance system based on Image Processing and Computer Vision using single or multiple camera mounted on car roof top facing the road, including LIDAR , RADAR sensor for detecting object, analysing road sourrounding and 3D modelling of the road environment. Also, in some works driver behaviour understanding using a camera facing the driver is used for drowriness, sleepiness, fatigue detection.
We have works on advanced driver assistance system, traffic safety,autonomous vehicle navigation and driver behaviour modelling using mutiple cameras, LIDAR, RADAR sensor etc. These works focus on using image processing and learning based method for lane detection, road segmentation, 3D modelling of road environment (e.g.[2],[4],[5],[6],[7],[21]). Parallax flow computation was used by Baehring et al. for detecting overtaking and close cutting vehicles [8]. For detecting and avoiding collison Radar, LIDAR, camera  and  omnidirectional camera was used in these works [12],[11],[13]. They focused on detecting using LIDAR sensor data classifying object as static and dynamic and tracking using extended Kalman filter and for getting a wide view of surrounding situation. For detection of forward collision Srinivasa et al. have used forward looking camera and radar data [9]. 

In some works driver inattentiveness was modelled using fatigue detection, drowsiness, eye tracking, cell phone usage etc. Trivedi et al modelled driver behavior using head movements for detecting driver gaze and distraction, targetting adavanced driver safety [20]. 
But works on localising vehicles with respect to roads and other cars have not done tll now, since knowing position of the car automatically can be great help for driver, so we have propsed this work.

\section{Localising Vehicle with respect to Lanes}
For localising vehicle on the road we estimate some related paramters like its current lane, shape of the road and its position from centerline. To compute this paramters we scan the road environment using wide angle camera sensor and extract lane markers. For lane detection we have proposed a novel method using Lab color space,2nd and 4th order steerable filters and improved Inverse Perspective Mapping. Below we describes our lane markers extraction algorithm.
\begin{figure}[h!]
  \centering
      \includegraphics[width=3.2in,height=2.5in]{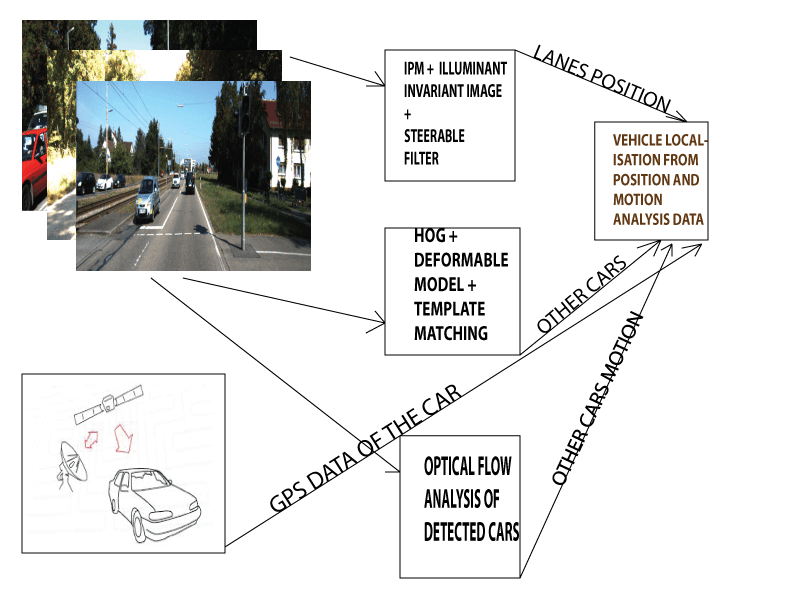}
\caption{Proposed Method}
\end{figure}

\subsection{Perspective effect}
In real world situation if we capture two parallel lines they appers to be converged to some distant points so their nature can't be understood in images. Road lanes are parallel ,includes both straight and curve roads, for detecting and localising them in images we need to remove the effect of perspective projection using Inverse Perspective Projection [1]. In this work we have presented a modified version of IPM instead of previous which is robust to a distance of 45m. No internal parameter calibration of camera is required for computation, which is a advantage over previous IPM implementation. Suppose Camera location with respect to car coordinates system (C{x},C{y},C{z}) where C{z} will be the height from ground lane 'h'. Optical axis make an angle $\theta$ known as pitch angle, $\gamma$ yaw angle and $\alpha$ as half of camera aperture as shown in fig.
\begin{figure}[h!]
  \centering
      \includegraphics[width=3.2in,height=2in]{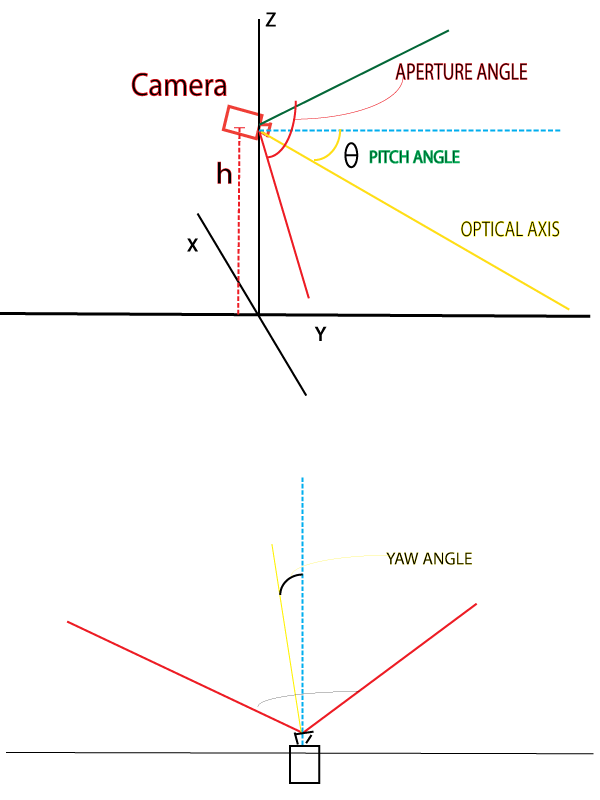}
\caption{Camera Setup, pitch angle, Yaw angle}
\end{figure}

To increase computational speed removing uninterested area from image we define horizon line from where our interested area will lie below "Horizon Limit" as shown in fig.

\begin{figure}[h!]
  \centering
      \includegraphics[width=3.2in,height=1.7in]{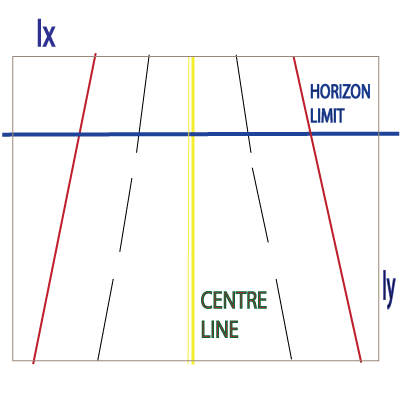}
\caption{Area of interest}
\end{figure}

 For derivation of IPM we will assume perfectly planar road. Image coordinates as (Ix,Iy) and real world coordinates as (x,y,0). If we suppose camera resolution as mxn, the we will get following mapping equation
\begin{equation}
\delta = tan^{-(m-1)/sqrt((m-1)^2 + (n-1)^2)*tan(\alpha)}
\end{equation}
\begin{equation}
\omega = tan^{-(n-1)/sqrt((m-1)^2 + (n-1)^2)*tan(\alpha)}
\end{equation}
\begin{equation}
hz = \frac{(m-1)*0.5}{(1-tan(\theta)/tan(\delta))} + 1
\end{equation}
\begin{equation}
x = h\frac{1+(1-2\frac{Ix-1}{m-1})tan(\delta)tan(\theta)}{tan\theta-(1-2\frac{Ix-1}{m-1})tan(\delta)}
\end{equation}

\begin{equation}
x = h\frac{(1-2\frac{Iy-1}{n-1})tan(\omega)}{sin\theta-(1-2\frac{Ix-1}{m-1})tan(\delta)cos\theta}
\end{equation}

where hz represent start row of image of interest.

\subsection{Feature Extraction}
For detecting vetical lines,2D steerable filters [7] are very effective to use,because of their seperability nature computation is faster than other filters. we have combined the result obtained from both 2nd and 4th order filters for extracting final lane markings on the basis of adaptive threshholding.

\begin{figure}[h!]
  \centering
      \includegraphics[width=3.2in,height=1.5in]{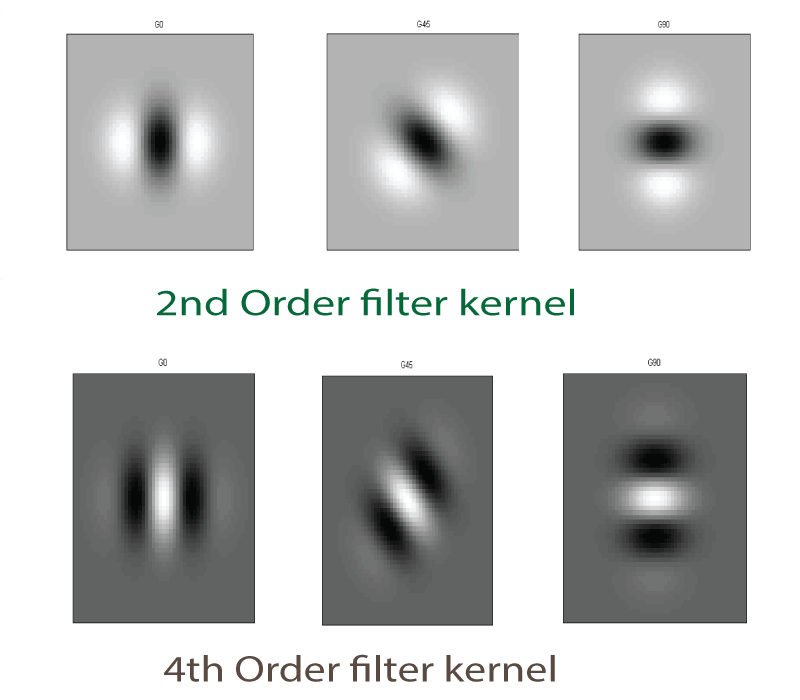}
\end{figure}

\begin{equation}
G_{2x}(x,y) = (\frac{4x^{2}}{\sigma^{4}} - \frac{2}{\sigma^{2}})e^{-\frac{x^{2} + y^{2}}{\sigma^{2}}}
\end{equation}
\begin{equation}
G_{xy}(x,y) = \frac{4xy}{\sigma^{4}}e^{-\frac{x^{2} + y^{2}}{\sigma^{2}}}
\end{equation}
\begin{equation}
G_{2y}(x,y) = (\frac{4y^{2}}{\sigma^{4}} - \frac{2}{\sigma^{2}})e^{-\frac{x^{2} + y^{2}}{\sigma^{2}}}
\end{equation}

\begin{equation}
G_{4x}(x,y) = (\frac{16x^{4}}{\sigma^{8}} - \frac{48x^{2}}{\sigma^{6}} - \frac{12}{\sigma^{4}})e^{-\frac{x^{2} + y^{2}}{\sigma^{2}}}
\end{equation}
\begin{equation}
G_{4y}(x,y) = (\frac{16y^{4}}{\sigma^{8}} - \frac{48y^{2}}{\sigma^{6}} - \frac{12}{\sigma^{4}})e^{-\frac{x^{2} + y^{2}}{\sigma^{2}}}
\end{equation}

\subsection{Cubic Interpolation and RANSAC}
Now for localising vehicle with respect to lanes we have used RANSAC [3] method for identifying and getting potential lane points position from extracted features points for fitting a parablic curves. Maximum of 8 curves can be identified by using our method. 
In most of the roads except center lane other lines are discontinuous, to get continuous edge to fit a polynomial in those plain areas cubic interpolation are very efficient. 
Our road model is given in equation (11), where $y_{0}$ is offset from vertical coordinate system and a,b,c are paramters.
\begin{equation}
Y = y_{0} + aX + bX^{2} + cX^{3}.
\end{equation}

\begin{figure}[h!]
  \centering
      \includegraphics[width=3.2in,height=1in]{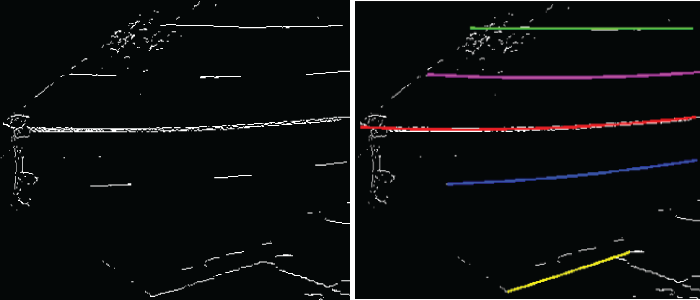}
\caption{Final feature points and line fitting after cubic interpolation}
\end{figure}
\subsection{Road Boundary Lane}
Only lane lines extraction can't give overall idea about car position if we don't know road boundary. Most of time road lane boundary are not so clear and even not paved mainly in indian situation, to cop up with this we need to get road area. For this we have used a 3 class based Gaussian mixture model for segmentation of the road region. Since IPM image's majority pixels are road part and cars and other obstacles present in the road area becomes noise in the IPM image, so GMM can be used efficiently for this task.Method is applied in illuminant invariant 45m accurate IPM image, this method perform efficiently for this purpose. 

Three custers used for segementation comprised of road region, sourrounding natural scenes and road obstacles. We have used predefined mean and covarinaces values for our clusters. For computation of these initial means and covariances we collected seperate pathes from train images for these three category and computed those values. This intialization gives us better result than random k means initializaton.
An iterative expectation maxmization based algorithm is used to compute final means, covariances and probability of each clusters in GMM.

At the end a bayesian classification techniques eq(12) is used to classify each pixels in image.
\begin{equation}
p(x/c_{i}) = p(c_{i})*e^{-\frac{(x - m_{c_{i}})^2}{2\sigma_{c_{i}}^2}}
\end{equation}
Here $c_{i}$ denotes a class i, $m_{c_{i}}$ means and $\sigma_{c_{i}}$ variances and prior probability $p(c_{i})$ of the class.

 Using vanishing point [5] estimation rest of the road boundary beyond 45m of images, which is not covered in IPM image can be apporximately modeled. 

\subsection{Lane Departure Angle}
To avoid potential risk of accident or misdriving, lane departure warning is very important. Using information from current position of vehicle with respect to lane specifically offset and optical flow computation this angle can be approximately computed.
\begin{equation}
\Lambda = \frac{v_{y}}{v_{x}}
\end{equation}

\begin{figure}[h!]
  \centering
      \includegraphics[width=0.5\textwidth]{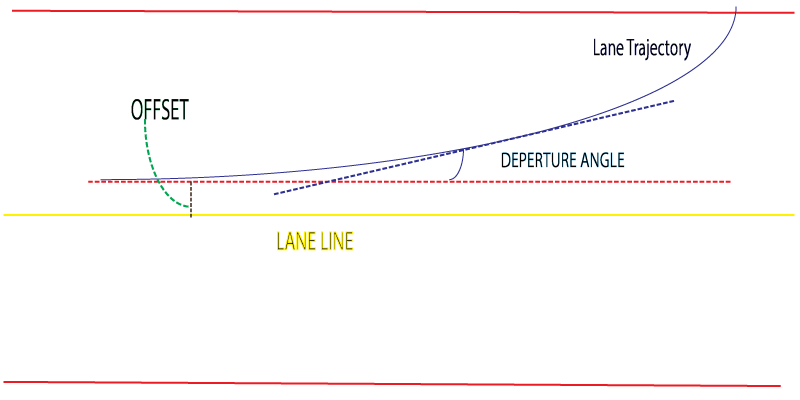}
\caption{Lane Departure angle }
\end{figure}

\section{Localising Vehicle with respect to Other cars}
For getting relative knowledge of the car locaion on the road, its location with respect to other car is estimated, like whether our car is behind from detected cars or left or right from those. This additional information will help as deciding factor for possible overtaking and getting more information about lane localisation. 

\subsection{Car detection for localisation on road}
Each frame of video obtained by our camera sensor is anlysised for detection of cars and other obstacles present in road.
For this purpose we have used Histogram of Gradient(HOG) [8] features based deformable part model(DPM), a very effective way to detect human and cars. HOG feature computation is based on gradient magnitude and angle. For descriptor computation, image is  divided into 8x8 cell, with 50\%  overlapping between nearby cells,then further divides the cell into 2x2 blocks for normalization of descriptor to make it illuminant invariant.
In case of DPM[15], each object is modeled as composition of its different parts. Training of this model is completely based on HOG feature and decomposition of object into its various parts and final model is obtained using latent SVM. Training phase produces root flter, corresponding to our car and part filters for representating various parts of car. This implementation used HOG pyramid based concept for better accuracy. At higher resolution HOG pyramid capture fine features and object can be detected accurately.This DPM based  method can detect car very efficiently under occlusion also. In Fig.7 we have shown model trained using [19], model is depicted using HOG feature representation picture and also its parts and result obtained by using this method.
\begin{figure}[h!]
  \centering
      \includegraphics[width=3.2in,height=4.3in]{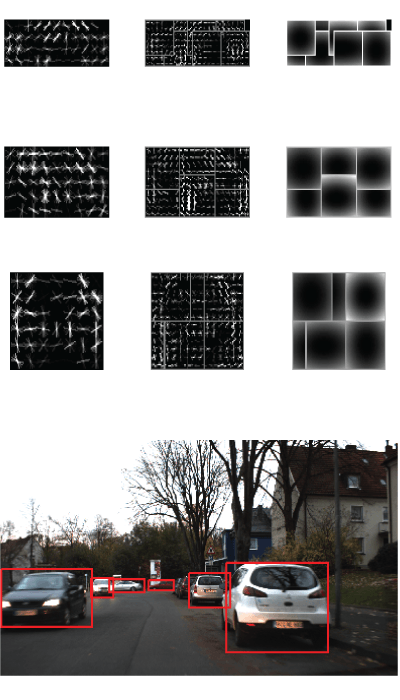}
\caption{Car model and detection result}
\end{figure}
Once we have detected other cars passing by, we wiil locate their current lane from their detected position on the image. After that we can confer about test vehicles position with respect to those cars.

\subsection{Enviroment Mobility Estimation}
After detecting a car, optical flow analysis can give its mobility estimation of those cars, which will provide a potential information about other cars movement. This estimation will also be useful for detecting possible traffic junction and trafic jam. Also our motivation to use optical flow lies mainly on very similar background involves in road environment.We have used optical flow computation described in this paper [16].

\subsection{GPS data Combination}

Since most of the time because of reflection from tall buliding in urban areas GPS data are not so accurate, to cop with this problem we will measure all the parameters as described above like vehicle current position with respect to road lane and other cars position. This parameters will give us local vehicles parameters with respect to road and global location from GPS data can give us approximate result about car location. This process can be extended to image based locality estimation by using specified training images.

\section{EXPERIMENT}
For showing the performance and reliablity of our algorithm in detecting lane, cars and road environment mobility, we have done a broader experimentation on 440x680 size images in different road condition. Our system was developed using MATLAB package in LINUX based OS with quad core intel i7 machine. For object detection part we have used voc-release library which is a state of art library for detecting object like cars pedestrians etc.
We have collected dataset in bangalore city road, with a wide angle camera sensor mounted on our test vehicle's roof pointing towards road at height 155cm from ground plane at speed of around 45km/h for testing our algorithm accuracy in indian condition, also for checking our algorithm for lane detection, we have used caltech [21] and KIT datatset [18]. This two daset contain image with different condition like, sunny road with shadow, urban road and highway etc.


\begin{figure*}

 \center

  \includegraphics[width=6.9in, height = 3.8in]{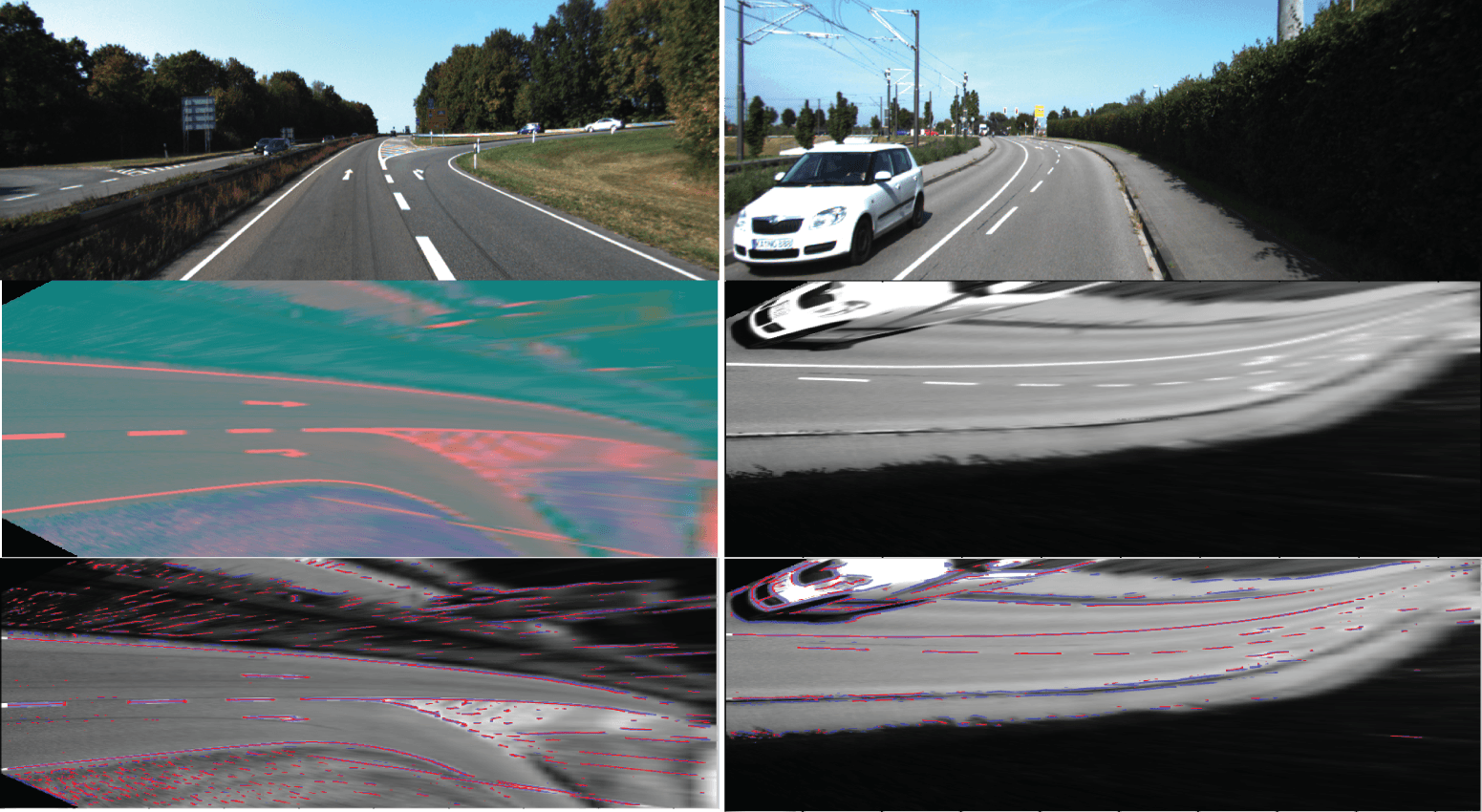}

  \caption{IPM image and detected possible lane features}

  \label{AAA}

\end{figure*}

\begin{figure*}

 \center

  \includegraphics[width=6.9in,height=2.8in]{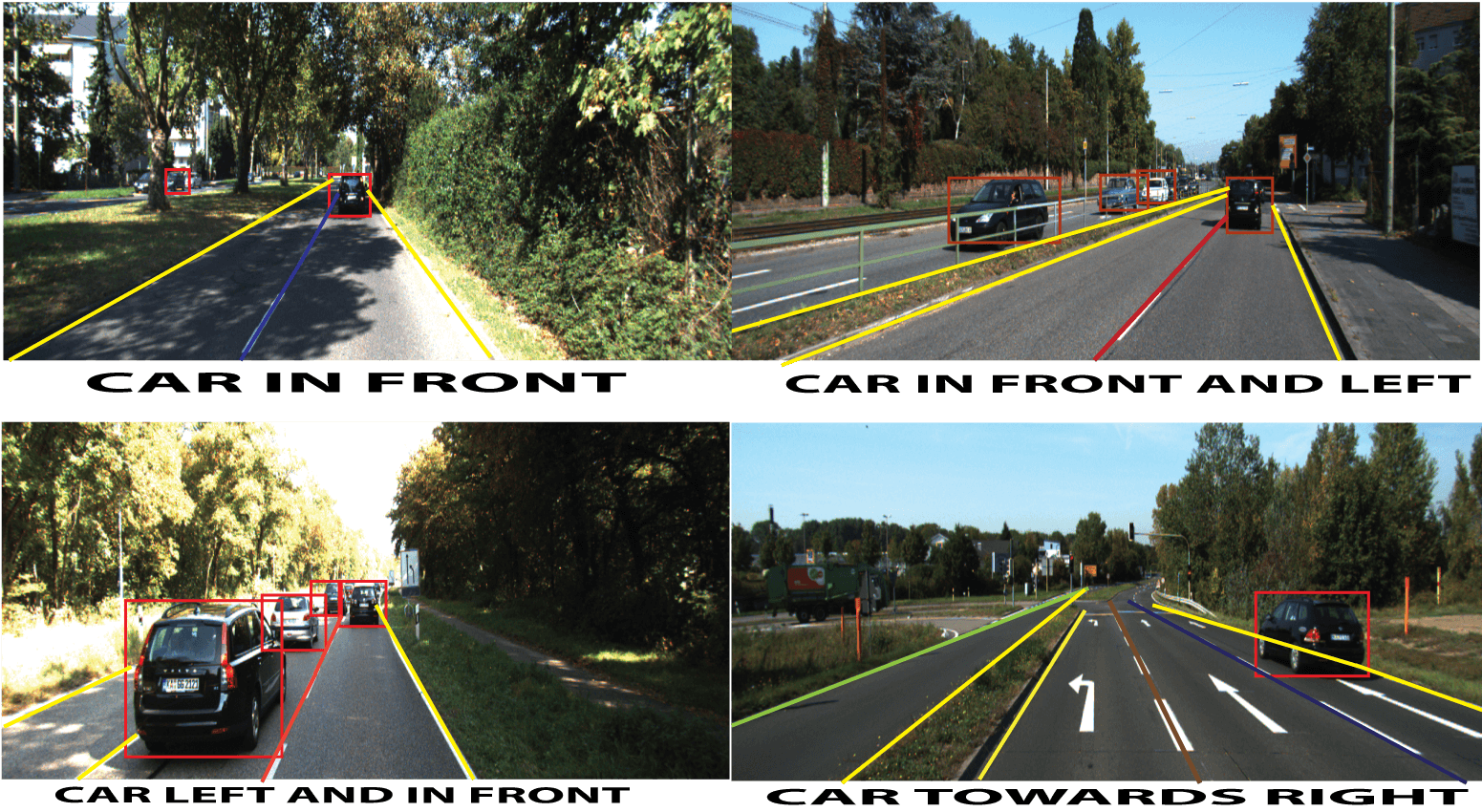}

  \caption{Result}

  \label{AAA}

\end{figure*}
With the combination of 2nd and 4th order seerable filters to detect edge in horizontal direction, result reduce extra outliers, which help in robust fitting of lane lines and better input to main RANSAC outliers removal and fitting lane lines. Some of the result is shown in Fig. [8] and Fig[9]. 
In case of some difficult road condition we were unable to detect lane lines in other side of two way roads, this situation is depicted in first row second column image in fig [9]. 

We are able to detect cars, other obstacles and was able to identify their relative location with respect our test setup, whether those cars are directly in front or in right side or in left side. This information of car location was used to analysis optical flow for getting better sense of their movements. If optical flow are stable then we can say that cars are moving at approximately same speed as our test cars and there are no other obstacles present. But if optical flows are changing rapidly we can confer about slow or high velocity of other cars with respect to our setup. 

We have obserevd that using illuminant invariance techniques using Lab color space gives better accuracy over normal RGB images for better detection of lane lines. 

\begin{table*}[t]
  \centering
  \begin{tabular}{*{20}{c}}
\hline
{\bf Database} & {\bf \#Frame} & {\bf \#detectedAll} & {\bf \#Boundary}  &{\bf CorrectRate} & {\bf False Positive} & {\bf CorrectBoundary}\\
\hline
KITTI & 600 & 565 & 591 & 94.26 \% & 6.79 \% & 98.44 \%\\  
\hline
Caltech & 1224 & 1189 & 1204 & 97.14 \% & 4.17\% & 98.36 \% \\  
\hline
Indian Road & 1200 & 1087 & 1131 & 90.58 \% & 12.37\% & 94.25 \%\\
\hline
\end{tabular} 
  \caption{CorrectRate of ego-lane evaluation(upto 45m) and Road Boundary Detection}
\end{table*}
In Table .[1] we have given analysis of accuracy obtained in lane detection and road boundary detection in three dataset. Lane detection and road boundary detection are building blocks for better accuracy of our method. It can be observed that this method perform very well in detection of the road lanes. Also we have observed that DPM car detection give very high accuracy, which enable us to locate vehicle position with respect to road. In addition to that GPS data accuracy determine method final accuracy to locate the car on the road. \\

It is not practical to give a quantitative analysis on vehicle local position estimation. Since number of cars in different roads are different and local vehicle position changes with respect to all this factors.


\section{Conclusion}
In this paper, a robust vehicle positioning system is presented using lane feature, car location and GPS data. This work demonstrated the posibility of using local position of vehicle with respect to road for better accuracy of vehicle position. The algorithm especially focus on enhancing safety in normal driving and for autonomous vehicles by keeping track of its local and global position. This specially usefull for urban areas with enormous amount of traffics to avoid from accident and driving safely. We have got considerable accuracy for localising vehicle with respect to lanes even in shadow and sunny road. This system include robust lane feature extaction using illuminant invariant techniques. In future we will develop a safe overtaking system.

\end{document}